\documentclass[10pt, a4paper]{article}

\usepackage[final]{lrec-coling2024} 
\usepackage{hyperref}
\usepackage{algorithm} 
\usepackage{algpseudocode}
\usepackage{amsmath}
\usepackage{multirow}
\usepackage{makecell}
\usepackage{tikz}
\usepackage{amssymb}
\usepackage{pifont}
\usepackage[euler]{textgreek}
\definecolor{myred}{RGB}{231 11 11}
\definecolor{myblue}{RGB}{0 198 237}
\definecolor{mygreen}{RGB}{117 185 79}
\definecolor{myyellow}{RGB}{250 190 0}
\definecolor{mypink}{RGB}{241 95 95}

\title{Explainable Multi-hop Question Generation: \\
An End-to-End Approach without Intermediate Question Labeling}

\name{Seonjeong Hwang$^1$, Yunsu Kim$^3$, Gary Geunbae Lee$^{1, 2}$} 

\address{$^1$Graduate School of Artificial Intelligence, POSTECH, Republic of Korea \\
         $^2$Department of Computer Science and Engineering, POSTECH, Republic of Korea  \\
         $^3$aiXplain, Inc. Los Gatos, CA, USA \\
         seonjeongh@postech.ac.kr, yunsu.kim@aixplain.com, gblee@postech.ac.kr\\}

\abstract{
In response to the increasing use of interactive artificial intelligence, the demand for the capacity to handle complex questions has increased.
Multi-hop question generation aims to generate complex questions that requires multi-step reasoning over several documents.
Previous studies have predominantly utilized end-to-end models, wherein questions are decoded based on the representation of context documents.
However, these approaches lack the ability to explain the reasoning process behind the generated multi-hop questions.
Additionally, the question rewriting approach, which incrementally increases the question complexity, also has limitations due to the requirement of labeling data for intermediate-stage questions.
In this paper, we introduce an end-to-end question rewriting model that increases question complexity through sequential rewriting. 
The proposed model has the advantage of training with only the final multi-hop questions, without intermediate questions.
Experimental results demonstrate the effectiveness of our model in generating complex questions, particularly 3- and 4-hop questions, which are appropriately paired with input answers.
We also prove that our model logically and incrementally increases the complexity of questions, and the generated multi-hop questions are also beneficial for training question answering models.
 \\ \newline \Keywords{Multi-hop Question Generation, Automatic Question Generation} }

\begin{document}

\maketitleabstract

\section{Introduction}

Question generation (QG) aims to generate questions related to the given context documents, with applications in various domains, including developing chatbots and educational tutoring systems and enhancing datasets used in question answering (QA) models. 
Recently, with advanced interactive artificial intelligence and chatbots, user questions have become complex, covering a wide range of information. 
Thus, the demand for machines capable of handling questions that necessitate intricate reasoning and contextual understanding has increased.

\begin{figure}[h]
\centering
\resizebox{0.95\columnwidth}{!}{%
\includegraphics{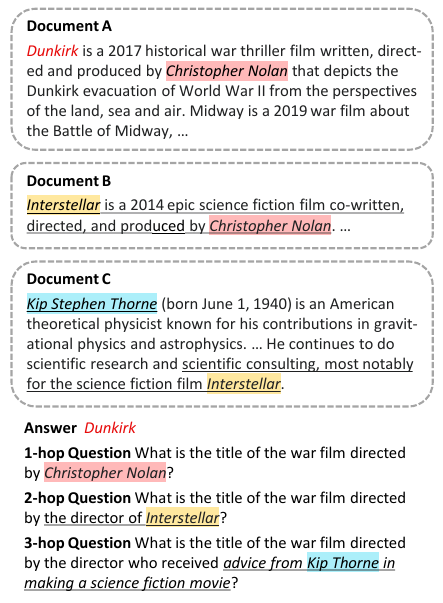}}
\caption{\label{fig:intro/data_example} Example of multi-hop question generation through question rewriting.}
\end{figure}

In early research, the focus was primarily on single-hop QG generating questions based on a single document or sentence \cite{du2017learning,zhao2018paragraph,sun2018answer,chan2019recurrent,alberti2019synthetic}. 
For example, the 1-hop question in Figure \ref{fig:intro/data_example} is constructed solely from the content of \textit{Document A}.
In real-world scenarios, however, we may not be able to recall direct information related to the question we want to ask.
Instead, we use additional information that indirectly describes the subject, as indicated in the 2- and 3-hop questions in the figure.

Multi-hop QG involves aggregating the related information scattered across multiple documents and generating questions that require multi-step reasoning.
In most previous work, graph-to-sequence (graph2seq) architectures are primarily used to extract meaningful relationships from several contexts and then generate multi-hop questions based on the context representation \cite{pan2020semantic, su2020multi, fei2022cqg}.
However, as the number of referenced documents increases, these end-to-end approaches face limitations in generating logically coherent multi-hop questions.
\citet{cheng2021guiding} proposed a step-by-step question rewriting framework to address this problem. 
This framework consists of a single-hop QG model and a question rewriting model, which increases the question complexity by leveraging additional information. 
To train the question rewriting model, the authors labeled intermediate questions that compose 2-hop questions.
Inspired by the method of \citet{cheng2021guiding}, we explore the step-by-step question rewriting approach, but it does not require intermediate question labeling.

In this paper, we introduce an End-to-End Question Rewriting (E2EQR) model, which initially generates single-hop questions and then sequentially rewrites them to increase the complexity. 
The E2EQR model is a type of recurrent neural network (RNN) with a sequence-to-sequence (seq2seq) Transformer \cite{vaswani2017attention} as its backbone.
In each step, the model takes a document and uses it to rewrite the question generated in the previous step.
Rather than using the generated question as the input for the subsequent steps, we implicitly transfer the hidden states computed during the decoding process in the prior steps, enabling end-to-end training without the ground truth for intermediate questions.
Additionally, we design a curriculum learning algorithm that allows the model to learn low- to high-complexity examples sequentially while preventing catastrophic forgetting for lower-hop questions.

In experiments, we conducted both quantitative and qualitative evaluations on the multi-hop questions generated by E2EQR.
According to the results, our method achieves performance comparable to the state-of-the-art model, even while performing the additional task of intermediate question generation.
Moreover, we prove the efficacy of the question rewriting approach in logically crafting complex questions that accurately align with the input answer.
We also observe that this precise capability in generating multi-hop QA pairs significantly contributes to data augmentation for training QA models.

Our contributions can be summarized as follows:
\begin{itemize}
    \item We introduce an explainable multi-hop QG model that performs step-by-step question rewriting to increase the question complexity\footnote{We release our code on \url{https://github.com/SeonjeongHwang/e2eQR}}.
    \item Our model is trained end-to-end without the need to label intermediate questions.
    \item Our question rewriting approach demonstrates effectiveness in logically generating complex questions that correspond to input answers, particularly in 3- and 4-hop QG scenarios.
    \item The synthetic multi-hop QA data also prove to be beneficial for training QA models.
\end{itemize}

\section{Related Work}

Early studies on QG have primarily focused on generating questions based on their syntactic and semantic patterns \cite{wolfe1976automatic,heilman2010good,heilman2011automatic,mazidi2014linguistic}.
However, these approaches were limited to generating factual questions from short sentences or paragraphs.
The seq2seq neural network has alleviated these limitations and enabled generating diverse types of questions. 
Recent work has proposed various approaches using attention-based seq2seq models \cite{du2017learning,zhao2018paragraph,sun2018answer} or pretrained language models \cite{chan2019recurrent,alberti2019synthetic}, which encode input documents and answers, and then generate related questions.
These studies aimed to generate single-hop questions using single-hop QA datasets, such as SQuAD  \cite{rajpurkar2016squad}, NewsQA \cite{trischler2016newsqa}, Natural Questions \cite{kwiatkowski2019natural}.

Multi-hop QG requires a comprehensive understanding of the content scattered across multiple documents.
\citet{pan2020semantic} and \citet{su2020multi} encoded input documents using graph neural networks (GNNs) to capture relationships between scattered information. 
\citet{fei2022cqg} proposed a hard-controlled generation framework that ensures multi-hop QG by forcing the decoding process to use key entities extracted from the input documents.
\citet{wang2020answer} and \citet{xie2020exploring} employed reinforcement learning to reflect the fluency, relevance or correspondence to the input answers of the generated questions for model training.
\citet{pan2021unsupervised} proposed an unsupervised method that generates single-hop questions and fuses them into multi-hop questions.
\citet{cheng2021guiding} proposed a step-by-step question rewriting framework to logically control the question difficulty.
The authors decomposed 2-hop questions into single-hop intermediate questions and trained two generation models: a single-hop QG model and a question rewriting model.
In this study, inspired by the approach of \citet{cheng2021guiding}, we develop an E2EQR model that can be trained without labeled intermediate questions.

\section{Method}

\begin{figure*}[h]
\centering
\resizebox{\textwidth}{!}{%
\includegraphics{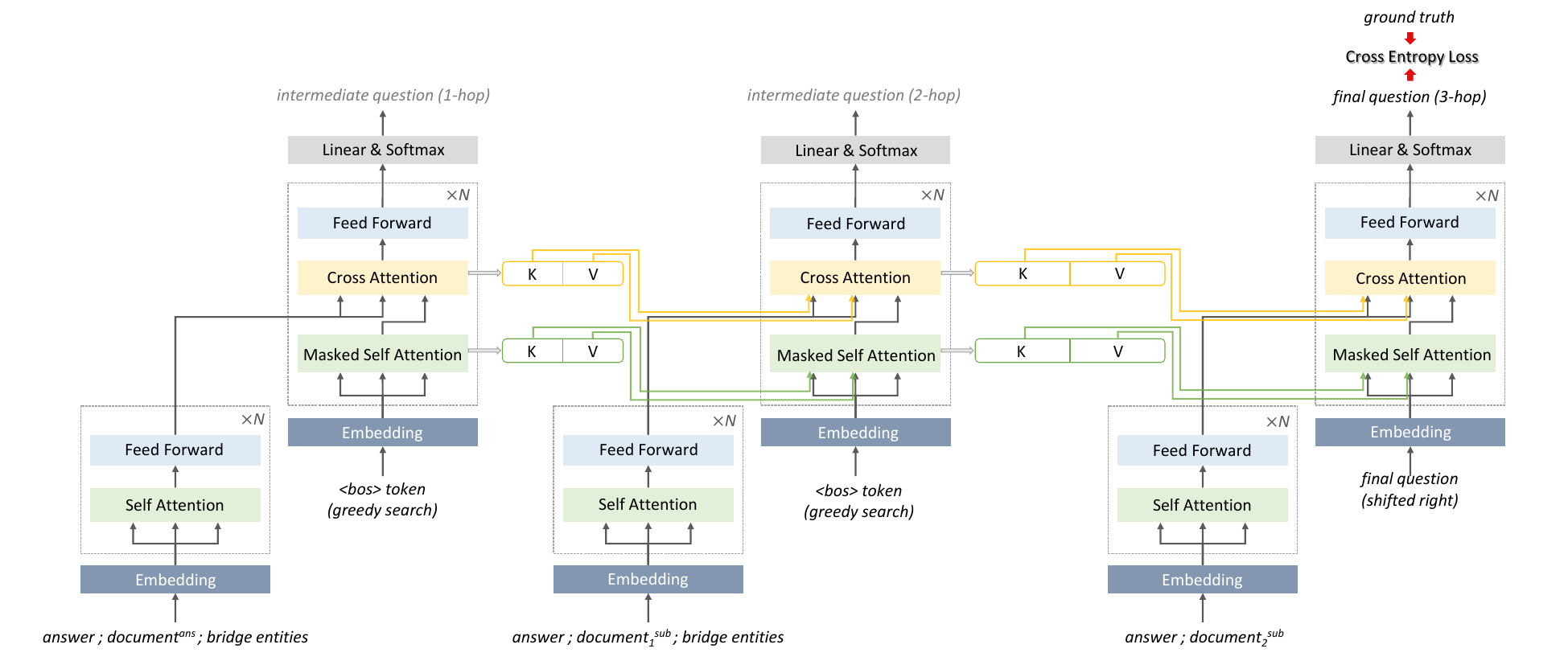}}
\caption{\label{fig:method/architecture} Unfolded architecture of the proposed model. Training process for 3-hop question generation. In the decoder, the multi-head masked self-attention and multi-head cross attention layers use the key and value matrices ($K$ and $V$) accumulated from the prior steps to rewrite the intermediate question generated in the previous step. We omitted the detailed elements of the Transformer \cite{vaswani2017attention} in this figure.}
\end{figure*}

The E2EQR model is a Transformer-based RNN that generates questions based on the current encoder input and hidden states computed while decoding in the previous steps.
As illustrated in Figure \ref{fig:method/architecture}, the model sequentially processes input elements and either generates the initial question or rewrites the previous one.
In the first step, the document that contains the answer is input into the model with the answer and bridge entities. 
Then, the model generates a 1-hop intermediate question using a greedy search.
The bridge entities carry the contents of future input documents, facilitating the generation of questions that can be rewritten.
In the $t$th step ($t > 1$), the model rewrites the $(t-1)$-hop question into the $t$-hop question.
Rather than explicitly taking the question generated in the previous step as input, our model relies on accumulated decoder hidden states to rewrite it.
This allows the model to be trained using only the ground truth of the final $\mathcal{N}$-hop questions.

\subsection{Problem Formulation}
Given the $\mathcal{N}$ document set $\{\mathcal{C}\}_{1:\mathcal{N}}$ and the target answer $\mathcal{A}$, the objective is to generate the $\mathcal{N}$-hop question $\mathcal{Q}^\mathcal{N}$ constructed based on inferences drawn from all given documents.
The document containing the answer and the rest of the documents are denoted as \textit{document}$^{ans}$ ($\mathcal{C}^{ans}$) and \textit{document}$^{sub}$ ($\mathcal{C}^{sub}$), respectively. 
The $t$th document $\mathcal{C}_t$ and $t$-hop question $\mathcal{Q}^t$ are the sequence of $l_t$ and $m_t$ tokens, respectively (i.e., $\mathcal{C}_t = c_{1:l_t}$ and $\mathcal{Q}^t = q_{1:m_t}$).

\subsection{\label{sec:arrangement} Document Arrangement and \,\,\,\,\,\,\,\,\,\,\,Bridge Entity Extraction}
In Figure \ref{fig:intro/data_example}, the entities \textit{``Christopher Nolan''} and \textit{``Interstellar''} are the bridges for \textit{Document} <\textit{A} to \textit{B}> and \textit{Document} <\textit{B} to \textit{C}>.
In addition, \textit{Document B} and \textit{C} are sequentially referenced to rewrite the 1-hop question as a 3-hop question.
As the example, the input documents should be input to the model in an order that allows sequential rewriting.
In addition, the model should know which entities will appear again in future input documents.
To extract bridge entities and arrange the input documents, we construct a document graph.

First, we extract bridge entities that commonly appear in more than two documents using named entity recognition and key phrase extraction tools\footnote{\url{https://demo.allennlp.org}}\footnote{\url{https://spacy.io}}.  
Next, we construct a document graph where the nodes represent input documents, and two nodes sharing bridge entities are connected by an edge.
Then we serialize the graph through a breadth-first-search with \textit{document}$^{ans}$ as a root.
The arranged documents are sequentially input to each step of E2EQR along with the answer and their bridge entities.

\subsection{\label{sec:E2EQR} End-to-End Question Rewriting}
Given $\mathcal{N}$ documents, bridge entities, and the answer, E2EQR aims to generate an optimal $\mathcal{N}$-hop question:
\begin{align*}
\mathcal{Q^N}^* = \arg \max_{\mathcal{Q^N}} P(\mathcal{Q^N} \mid \{ \mathcal{C} \} _{1:\mathcal{N}}, \{ \mathcal{B} \}_{1:(\mathcal{N}-1)}, \mathcal{A} ; \theta),
\end{align*}
where $\mathcal{B}_t$ represents a set of bridge entities input with $\mathcal{C}_t$, and $\theta$ denotes the model parameters, consisting of the encoder and decoder parameters: $\theta_{enc}$ and $\theta_{dec}$. 
The model does not take the bridge entities in the $\mathcal{N}$th step, and the model parameters are shared in all steps.

In the first step, the encoder takes the input sequence and returns the output representation $H_1 \in R^{l_1 \times dim_{model}}$:
\begin{align*}
H_1 = \mathrm{Encoder}(\mathcal{C}_1, \mathcal{B}_1, \mathcal{A} ; \theta_{enc}),
\end{align*}
where $l_1$ represents the length of the input sequence, and $dim_{model}$ denotes the output dimension.
With the encoder output representation, the decoder performs autoregressive decoding from the \textit{<bos>} token until the \textit{<eos>} token is returned:
\begin{align*}
P(\mathcal{Q}^1 \mid H_1; \theta_{dec}) = \prod_{i=1}^{m_1}P(q_i \mid q_{1:i-1}, H_1;\theta_{dec}),
\end{align*}
where the length of the 1-hop question is $m_1$.

In the $t$th ($t > 1$) step, the encoder output representation $H_t \in R^{l_t \times dim_{model}}$ is obtained in the same way as the first step. 
However, from the $t$th step, the decoder uses the encoder output representation and the decoder hidden states calculated in the prior steps.

The Transformer decoder contains the multi-head masked self-attention (SA) layers and multi-head cross-attention (CA) layers\footnote{For a simple description, we omitted concepts regarding Transformer multi-head attention.}, and the layer output is the attention value computed from query $Q$, key $K$, and value $V$ matrices \cite{vaswani2017attention}.
The SA mechanism employs the projections of the decoder input representation as $Q \in R^{m_t^\prime \times d_k}$, $K \in R^{m_t^\prime \times d_k}$, and $V \in R^{m_t^\prime \times d_k}$ matrices, where $m_t^\prime$ ($1 \leq m_t^\prime \leq m_t$) is the decoder input length, and $d_k$ is the attention hidden dimension.
In the CA layer, the key and value matrices are the projections of the encoder output representation $H_t$ (i.e., $K \in R^{l_t \times d_k}$ and $V \in R^{l_t \times d_k}$).

In the proposed method, we introduce the \textbf{accumulated SA} and \textbf{accumulated CA} mechanisms, which employ the accumulated key and value matrices, $K_{1:t}$ and $Q_{1:t}$, the concatenations of matrices computed in the prior steps and the current $t$th step.
In other words, the scaled dot-product attention function in the accumulated attention layers is as follows:
\begin{align*}
\mathrm{Attention}(Q_t, K_{1:t}, V_{1:t}) = \mathrm{Softmax}(\frac{Q_t K_{1:t}^T}{\sqrt{d_k}})V_{1:t}.
\end{align*}
The size of the matrices $K_{1:t}$ and $V_{1:t}$ is $(\sum_{i=1}^{i=t-1} m_i + m_t^\prime) \times d_k$ in the accumulated SA and  $(\sum_{i=1}^{i=t} l_t) \times d_k$ in the accumulated CA.
In summary, the question decoding in the $t$th step is as follows:
\begin{align*}
& P(\mathcal{Q}^t \mid H_t, K_{1:t-1}, V_{1:t-1}; \theta_{dec}) \\
& = \prod_{i=1}^{m_t}P(q_i \mid q_{1:i-1}, H_t, K_{1:t-1}, V_{1:t-1};\theta_{dec}).
\end{align*}

The accumulated SA and CA mechanisms allow the use of the information from the intermediate questions and input documents of the prior steps, respectively.
Moreover, instead of using the question generated in the previous step as input to the next step, our model performs implicit question rewriting via these methods.
Consequently, this method allows end-to-end training of the model without the ground truth of the intermediate questions.
We train our model using the cross-entropy loss for the question generated in the final step.

\subsection{\label{sec:curriculum} Adaptive Curriculum Learning}
The proposed model generates $\mathcal{N}$-hop questions based on the capability of generating 1- to ($\mathcal{N}-1$)-hop questions.
Therefore, we design a curriculum learning algorithm where the model learns sequentially from low-hop questions to high-hop questions.
However, we empirically observed that sequential learning by complexity level causes catastrophic forgetting.
Thus, we apply intensive multitask learning, where the weight assigned to each complexity changes as the training progresses.

\begin{algorithm}[h]
\small
\renewcommand{\algorithmicrequire}{\textbf{Input:}}
\caption{Adaptive Curriculum Learning}\label{algo:curriculum learning}
\begin{algorithmic}[1]
\Require $\{ D_1, D_2, \cdot \cdot \cdot, D_\mathcal{N} \}$, E2EQR$_\theta$, $\alpha$, $\gamma_{low}$, $\gamma_{high}$, $\rho$
\For{$\mathcal{H}=1$ to $\mathcal{N}$}
    \State Training examples $\mathfrak{D} \leftarrow D_1 \cup \cdot \cdot \cdot \cup D_\mathcal{H}$
    \For{$h=\mathcal{H}+1$ to $\mathcal{N}$}
        \State $\mathfrak{D} \leftarrow \mathfrak{D} \cup D_h^{part}$, where $n(D_h^{part}) = \rho \cdot n(D_h)$
    \EndFor
    \For{$i=1$ to $n(\mathfrak{D})$}
        \State $x_i, y_i \leftarrow \mathfrak{D}_i$
        \State $\mathcal{L}_i \leftarrow $ \textit{CrossEntropyLoss}($y_i$, E2EQR$_{\theta}$($x_i$))
        \If{$i < \mathcal{H}$}
            \State $\mathcal{L}_i \leftarrow \gamma_{low} \cdot \mathcal{L}_i$
        \EndIf
        \If{$i > \mathcal{H}$}
            \State $\mathcal{L}_i \leftarrow \gamma_{high} \cdot \mathcal{L}_i$
        \EndIf
    \EndFor
    \State $\mathcal{L} \leftarrow \sum_{i=1}^{n(\mathfrak{D})} \mathcal{L}_i / n(\mathfrak{D})$
    \State $\theta \leftarrow \theta - \alpha \cdot \frac{\partial \mathcal{L}}{\partial \theta}$
\EndFor
\end{algorithmic}
\end{algorithm}

Algorithm \ref{algo:curriculum learning} describes the overall process of training E2EQR$_\theta$ with examples $\{ D_1, D_2, \cdot \cdot \cdot, D_\mathcal{N} \}$ grouped by the question complexity.
In this algorithm, the main complexity $\mathcal{H}$ is assigned to each iteration, which increases from 1 to $\mathcal{N}$, and the rest are treated as subcomplexities (Line 1).
In the training iteration with the main complexity $\mathcal{H}$, all examples from $D_1$ to $D_\mathcal{H}$ are used as training examples $\mathfrak{D}$ (Line 2).
The examples from $D_1$ to $D_{\mathcal{H}-1}$ are used to prevent catastrophic forgetting.
In addition, examples $D_h^{part}$ with a higher complexity than $\mathcal{H}$ are also used to prevent overfitting for lower-hop questions, and the number of examples is controlled using the suppression ratio $\rho$ (Lines 3--5).

Lines 6--15 display the process of updating model parameters $\theta$ using the examples $\mathfrak{D}$ and the learning rate $\alpha$.
At this stage, we employ two hyperparameters: the loss weight for lower complexity levels $\gamma_{low}$ and the loss weight for higher complexity levels $\gamma_{high}$.
These loss weights allows the model to focus on training examples of the main complexity.
We use different loss weights for the two groups because the model already learns the examples of lower complexities.
The loss weight is applied to the computed loss according to the data complexity.
This algorithm represents a simplified version, and in actual experiments, we switched the main complexity after optimizing the model for that specific complexity.
In addition, we used mini-batch stochastic gradient descent and learning rate scheduling, which linearly decreases the learning rate.

\section{Experiments}

\subsection{Dataset}
We used two multi-hop QA datasets, MuSiQue \citelanguageresource{trivedi2022musique} and HotpotQA \citelanguageresource{yang2018hotpotqa}, to compare our model with the baselines.
HotpotQA contains crowd-sourced 1- and 2-hop questions that require reasoning on supporting facts collected from Wikipedia.
MuSiQue is a more challenging dataset, containing up to 4-hop questions constructed by compositing 1-hop questions.
The authors of this dataset aimed to address certain shortcomings in HotpotQA, where answers can be derived by reasoning from only partial supporting evidence.
In the experiments, we mainly used MusiQue to verify that our model robustly generates questions of varying complexity levels.

The test sets for both datasets are not publicly available; thus, we used the examples from the part of examples from the original training set as the validation set and the original validation set as the test set.
Therefore, we used 25,483/700/2,417 examples and 89,947/500/7,405 examples for training, validation and test on MuSiQue and HotpotQA, respectively.
Additionally, we split the MuSiQue training set into 12,742 $seen$ and 12,741 $unseen$ examples for generating a synthetic training set in Section \ref{sec:augmentation}.

\subsection{Metric}
We measured BLEU \cite{papineni2002bleu}, METEOR \cite{banerjee2005meteor}, and ROUGE \cite{lin2004rouge} scores using \texttt{pycocoevalcap}\footnote{\url{https://github.com/salaniz/pycocoevalcap}}.
All metrics evaluate the n-gram similarity between the prediction and ground truth.
The BLEU score measures n-gram word precision and we specifically used the 4-gram-based BLEU (BLEU-4).
The METEOR score calculates the F1 score for explicit word-to-word alignment, and ROUGE evaluates the n-gram similarity using the F1 score.
We used ROUGE-L, focusing on the longest common subsequence.
However, it is insufficient to evaluate the question quality based only on the similarities to reference questions.
Therefore, we conducted a qualitative analysis through human evaluation.

\subsection{Baselines}
We compared our model with following strong multi-hop QG baselines\footnote{We used GitHub source codes released by the authors to train the baselines on MuSiQue.}:
\begin{itemize}
    \item \textbf{DP-Graph} \cite{pan2020semantic} is a graph2seq model that generates questions based on the word-level document representation and node-level semantic graph representation encoded using an attention-based GNN.
    \item \textbf{CQG} \cite{fei2022cqg} generates multi-hop questions using a hard-controlled generator, which controls the decoding process so that the key entities extracted from documents are included in the generated question.
    \item \textbf{MulQG} \cite{su2020multi} is a graph2seq model using a graph convolution network to encode the input documents in consideration of the answer.
    \item \textbf{BART} \cite{lewis2020bart} is a seq2seq Transformer pretrained with the text denoising task. We used the \textit{large} model and trained the model to generate multi-hop questions given the concatenation of the answer and documents.
\end{itemize}
We also examined whether a large language model can perform multi-hop QG through in-context learning.
The results are reported in Appendix \ref{sec:appendix/llm}.

\begin{table*}[h]
\centering
\renewcommand*{\arraystretch}{1.2}
\resizebox{\textwidth}{!}{
\begin{tabular}{l|ccc|ccc|ccc|c}
\Xhline{1.2pt}
\multirow{2}{*}{Model} & \multicolumn{3}{c|}{2-hop} & \multicolumn{3}{c|}{3-hop} & \multicolumn{3}{c|}{4-hop} & \multirow{2}{*}{\begin{tabular}[c]{@{}c@{}}Intermediate\\ Question\end{tabular}} \\ \cline{2-10}
                          & BLEU-4    & METEOR  & ROUGE-L  & BLEU-4    & METEOR  & ROUGE-L  & BLEU-4    & METEOR  & ROUGE-L  &                              \\ \hline
DP-Graph \cite{pan2020semantic}     & 5.14    & 10.80    & 28.69  & 4.33    & 10.37   & 28.33  & 4.47    & 9.87    & 28.01  &                              \\
CQG \cite{fei2022cqg}     & 9.64    & 16.20   & 33.98  & 7.79    & 13.77   & 31.12  & 5.14    & 11.93   & 26.48  &                              \\
MulQG \cite{su2020multi}  & 9.56    & 15.41   & 37.18  & 9.23    & 14.35   & 35.66  & 7.13    & 12.43   & 31.88  &                              \\
BART \cite{lewis2020bart} & 20.84   & 25.91   & 43.81  & 17.64   & 22.93   & 41.16  & 16.11   & 20.63   & 37.02  &                              \\
E2EQR   & 20.33   & 25.64   & 44.01  & 17.02    & 22.33   & 40.04  & 15.34   & 19.78   & 36.98  & \checkmark                   \\ \Xhline{1.2pt} 
\end{tabular}}
\caption{\label{tab:results/automatic} Automatic evaluation results on the MusiQue test set.}
\end{table*}

\begin{table*}[h]
\centering
\renewcommand*{\arraystretch}{1.2}
\resizebox{\textwidth}{!}{
\begin{tabular}{l|ccc|ccc|ccc}
\Xhline{1pt}
\multirow{2}{*}{Model} & \multicolumn{3}{c|}{2-hop} & \multicolumn{3}{c|}{3-hop} & \multicolumn{3}{c}{4-hop} \\ \cline{2-10} 
                       & Fluency  & Complexity ($\leq 2$)  & Answer Matching      & Fluency   & Complexity ($\leq 3$)  & Answer Matching     & Fluency  & Complexity ($\leq 4$) & Answer Matching     \\ \hline
BART                   & 4.80     & 1.93                   & 87.5\%               & 4.88      & 2.54                   & 75.0\%              & 4.78     & 2.92                  & 73.8\%              \\
E2EQR                  & 4.92     & 1.99                   & 87.5\%               & 4.83      & 2.50                   & 82.5\%              & 4.60     & 2.96                  & 80.0\%              \\ \hline
Ground Truth           & 4.85     & 1.99                   & 95.0\%               & 4.85      & 2.65                   & 78.8\%              & 4.65     & 3.29                  & 77.5\%              \\ \Xhline{1pt}
\end{tabular}}
\caption{\label{tab:results/human} Human evaluation of questions generated by multi-hop QG models and the ground truth.}
\end{table*}

\subsection{Implementation Details}
We initialized our model parameters using \texttt{facebook/BART-large} released on Hugging Face\footnote{\url{https://huggingface.co}}.
We used AdamW \cite{loshchilov2017decoupled} optimizer with a batch size of 8, a learning rate of 3e-5, 1000 learning rate warm-up steps.
The hyperparameters $\rho$, $\gamma_{low}$, and $\gamma_{high}$ were set to 0.1, 0.8, and 0.1, which is the optimal combination on the validation set and searched from \{1, 0.1, 0.01\}, \{1, 0.8, 0.5\} and \{0.1, 0.01\}, respectively.
Our models and baselines were trained using five random seeds and we reported the average performance in Section \ref{sec:automatic}.

\section{Results}

\subsection{\label{sec:automatic} Automatic Evaluation}
Table \ref{tab:results/automatic} summarizes the performance of E2EQR and baseline models on the MuSiQue test set.
According to the results, our model significantly surpasses DP-Graph, CQG, and MulQG across data of all complexity levels.
These models are optimized for the HotpotQA dataset, which comprises 2-hop questions with shorter supporting evidence compared to MuSiQue.
We attribute the poor performance of these models on MuSiQue to this disparity in data characteristics.
E2EQR and BART exhibit comparable results, as they share the same backbone.
However, our model, despite having the same number of parameters as BART, generates multi-hop questions as well as their intermediate questions.
We also report the results on HotpotQA in Appendix \ref{sec:appendix/hotpotqa}.

\subsection{Human Evaluation}
We conducted further investigation into the quality of synthetic multi-hop questions with the assistance of native English speakers\footnote{We enlisted human evaluators from Amazon Mechanical Turk (\url{https://www.mturk.com}) and Upwork (\url{https://www.upwork.com}).}.
We compared the questions generated by E2EQR and the strongest baseline model, BART, against ground-truth questions.
We randomly selected 40 document sets for each complexity level, and the three different questions based on each document set were evaluated according to three criteria:
\textit{Fluency} assesses whether the question is logically constructed and adheres to correct grammar. Each question is assigned a score of 1 (poor), 3 (acceptable), or 5 (good).
\textit{Complexity} indicates the number of documents relevant to the question. The raters list the indices of any documents that are referenced to understand the question, and we report the average number of selected documents.
\textit{Answer Matching} determines whether the question asks about the input answer.

Table \ref{tab:results/human} presents the average scores rated by two evaluators.
In terms of fluency, the questions generated by both multi-hop QG models are assessed as fluent and grammatically sound, akin to ground-truth questions.
Notably, E2EQR demonstrates superior performance over BART in answer matching criteria in 3- and 4-hop QG, despite both models receiving similar complexity scores.
The BART model is trained using end-to-end teacher forcing with ground-truth questions.
Consequently, the model's capacity to discern relationships among input documents and generate questions pertinent to the input answer is constrained.
Conversely, our model increases question complexity based on the documents and bridge entities entered at each rewriting step, thereby generating questions consistent with the input answer.
Given that our model produces appropriate multi-hop QA pairs, it also proves effective in augmenting data for multi-hop QA in Section \ref{sec:augmentation}.

\begin{table}[h]
\centering
\small
\renewcommand*{\arraystretch}{1.2}
\begin{tabular}{ccc}
\Xhline{1pt}
Fluency & Complexity Growth  & Relevance  \\ \hline
4.60    & 87.9\%             & 86.8\%     \\ \Xhline{1pt}
\end{tabular}
\caption{\label{tab:results/intermediate} Human evaluation of the question rewriting in our model.}
\end{table}

We additionally verified whether E2EQR successfully rewrites the intermediate question generated in the preceding step, taking into account the newly entered document.
Among the final and intermediate questions generated by E2EQR, 100 pairs of questions, before and after rewriting, were randomly selected. 
The rewritten questions were then evaluated by two English native speakers based on three criteria: \textit{Fluency}, \textit{Complexity Growth}, and \textit{Relevance}.
Complexity growth assesses whether the rewritten question exhibits greater complexity than its predecessor.
Relevance indicates whether the question rewriting process references the content of the input document.

As shown in Table \ref{tab:results/intermediate}, the intermediate questions generated by E2EQR display both logical and grammatical correctness.
Additionally, over three-quarters of the questions were successfully rewritten based on the provided documents, resulting in greater complexity compared to their predecessors.
Generating intermediate questions enables the model to logically enhance the complexity of the question and facilitate better understanding of the results.
Furthermore, multi-hop questions feature intricate sentence structures, posing challenges for error detection.
Therefore, leveraging intermediate questions enables identification and filtering of incorrect multi-hop questions at an early stage.

\section{Analysis and Discussion}

\subsection{\label{sec:augmentation} Data Augmentation for Question Answering}
We assess the efficacy of synthetic data generated by E2EQR for training multi-hop QA models.
First, we trained the models using the training$_{seen}$ set of MuSiQue.
The models then generated questions from unseen documents in the training$_{unseen}$ set.

\begin{figure}[h]
\centering
\resizebox{\columnwidth}{!}{%
\includegraphics{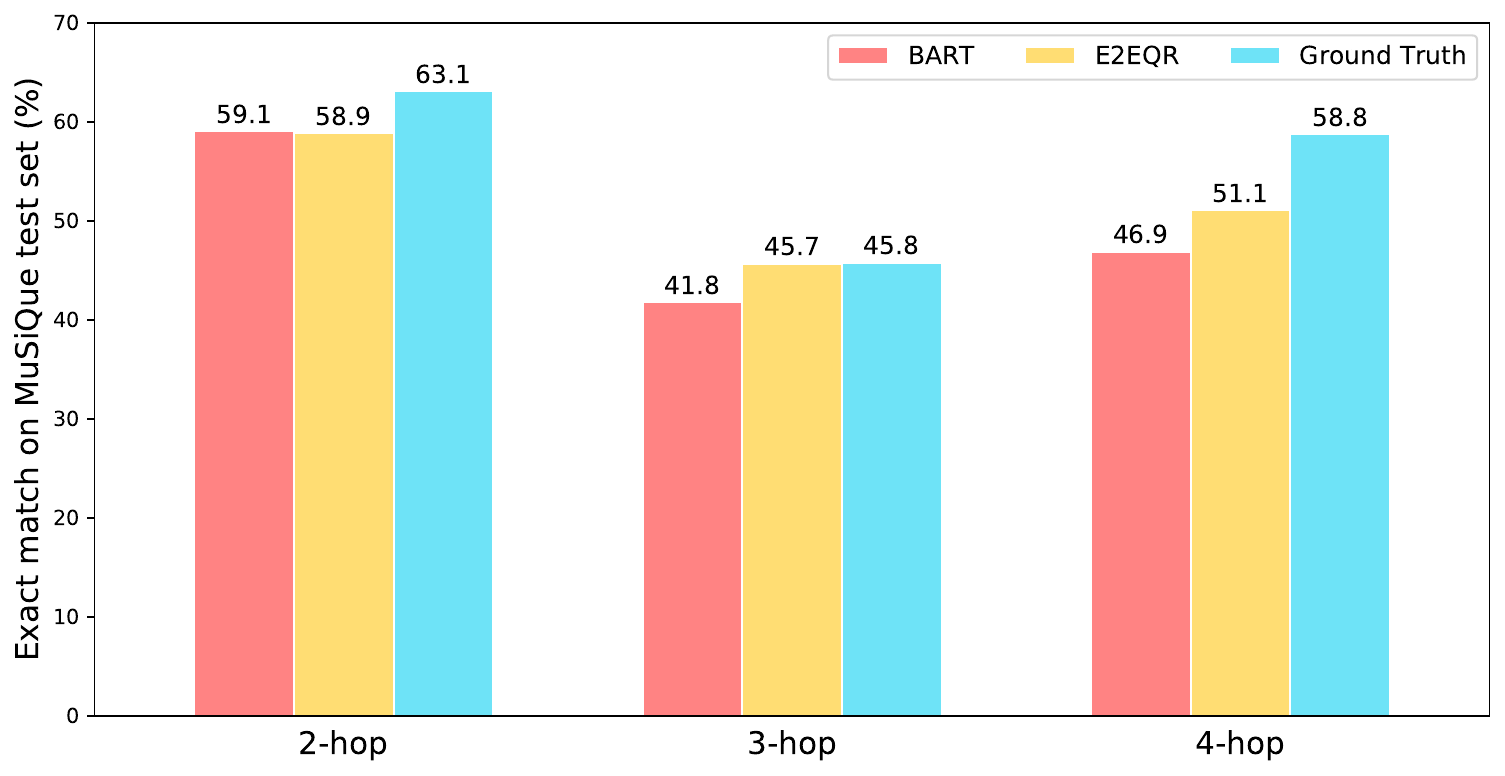}}
\caption{\label{fig:results/qa_results} The performance of three QA models trained on the synthetic data generated by BART and E2EQR and the MuSiQue training$_{unseen}$ set (Ground Truth), respectively.}
\end{figure}

Figure \ref{fig:results/qa_results} describes the best performance of QA models trained on the synthetic data of E2EQR and BART, and the ground truth, respectively.
We implemented the QA models based on T5 \cite{raffel2020exploring}.
According to the results, the QA model trained on our data achieves 3.9 and 4.2 higher performance in 3- and 4-hop examples, respectively, than the model trained on the data of BART.
In the 2-hop examples, both QG models have the similar effects in training QA models.

\subsection{Ablation Study}

\begin{table}[h]
\centering
\renewcommand*{\arraystretch}{1.2}
\resizebox{\columnwidth}{!}{%
\begin{tabular}{l|ccc}
\Xhline{1pt}
Model                     & 2-hop & 3-hop & 4-hop \\ \hline
E2EQR                     & 44.61 & 40.31 & 38.41 \\
E2EQR w/o Accumulated SA  & 43.32 & 39.94 & 35.14 \\ 
E2EQR w/o Accumulated CA  & 41.84 & 36.69 & 35.35 \\ \Xhline{1pt}
\end{tabular}}
\caption{\label{tab:results/ablation} The ablation study for E2EQR. The best ROUGE-L score is reported.}
\end{table}

We further investigated the effects of the accumulated SA and CA mechanisms on the performance of E2EQR.
Table \ref{tab:results/ablation} shows the performance of E2EQR trained without each component.
For each case, we used the original SA or CA mechanism instead of the accumulated ones.
Without the accumulated SA, a performance degradation of 1.29, 0.37 and 3.27 points is observed in 2-, 3- and 4-hop QG, respectively.
In addition, the model without the accumulated CA mechanism has a more significant performance drop, especially in 2- and 3-hop QG.
Consequently, we prove that E2EQR rewrites the questions relying on the representations of the previous input documents and intermediate questions, which are transferred through the accumulated CA and SA.

\subsection{Analysis of Training Strategies}

\begin{table}[h]
\centering
\renewcommand*{\arraystretch}{1.2}
\resizebox{\columnwidth}{!}{%
\begin{tabular}{ll|ccc}
\Xhline{1pt}
\multicolumn{2}{l|}{Training Strategy}                       & 2-hop & 3-hop & 4-hop \\ \hline
\multicolumn{2}{l|}{1. Standard Training}               & 43.00 & 40.99 & 34.03 \\ \hline
\multirow{3}{*}{Curriculum Learning} & 2. Step-by-step & 28.87 & 38.69 & 33.95 \\
                                     & 3. Cumulative  & 43.67 & 38.99 & 36.66 \\
                                     & 4. Adaptive \textit{(ours)}   & 44.61 & 40.31 & 38.41 \\ \Xhline{1pt}
\end{tabular}}
\caption{\label{tab:results/traning_ablation} Performance of E2EQR trained using different strategies.}
\end{table}

\begin{table*}[h]
\centering
\scriptsize
\renewcommand*{\arraystretch}{1.2}
\begin{tabular}{m{15.5cm}}
\Xhline{1pt}
\textit{Example 1} \\ \hline
\textbf{[Document A] African-American candidates for President of the United States} \\
In 1888 Frederick Douglass was invited to speak at the Republican National Convention. Afterward during the roll call vote, he received one vote, so was nominally a candidate for the presidency. In those years, the candidates for the presidency and vice presidency were chosen by state representatives voting at the nominating convention. Many decisions were made by negotiations of state and party leaders behind closed doors. Douglass was not a serious candidate in contemporary terms. \\ 
\textbf{[Document B] Helen Pitts Douglass} \\
Helen Pitts Douglass (1838--1903) was an American suffragist and abolitionist, known for being the second wife of Frederick Douglass. She also created the Frederick Douglass Memorial and Historical Association. \\ \hline

\textbf{Answer: } Helen Pitts Douglass \\
\hline
\textbf{DP-Graph: } who was the child of the person responsible for the political party with douglass reid ? \\
\textbf{CQG: } who was the spouse of the president of the united states as a member? \\
\textbf{MulQG: } who are the two leaders of the opposition in the province where douglass is located ? \\
\textbf{BART: } Who is the spouse of the person who was invited to speak at the Republican National Convention in 1888? \\
\textbf{E2EQR: } Who is the spouse of the person who was invited to speak at the 1888 Republican convention? \\
\textbf{Ground Truth: } Who is the spouse of the first nominated African American presidential candidate? \\
\Xhline{1pt}

\textit{Example 2} \\ \hline
\textbf{[Document A] Rio Linda High School} \\
Rio Linda High School is a high school located in Rio Linda, Sacramento, CA. It has an enrollment of 2,035 students. It is part of the Twin Rivers Unified School District, and was formerly part of the Grant Unified School District. \\ 

\textbf{[Document B] Area code 951} \\
Area code 951 is a California telephone area code that was split from area code 909 on July 17, 2004. It covers western Riverside County, including, Beaumont, Corona, Canyon Lake, Riverside, Temescal Canyon, ... \\

\textbf{[Document C] History of Sacramento, California} \\
The history of Sacramento, California, began with its founding by Samuel Brannan and John Augustus Sutter, Jr. in 1848 around an embarcadero that his father, John Sutter, Sr. constructed at the confluence of the American and Sacramento Rivers a few years prior. \\

\textbf{[Document D] California Gold Rush} \\
Rumors of the discovery of gold were confirmed in March 1848 by San Francisco newspaper publisher and merchant Samuel Brannan. Brannan hurriedly set up a store to sell gold prospecting supplies, ... \\ \hline

\textbf{Answer: } Rio Linda \\ \hline
\textbf{DP-Graph: } what city is the capital of the county where the town of linda is headquartered ? \\
\textbf{CQG: } what is the administrative territorial entity where the gold rush is located? \\
\textbf{MulQG: } what is the area code for the state where area code 951 is located ? \\
\textbf{BART: } What is the capital of the area code 951 of the state where the California Gold Rush began? \\
\textbf{E2EQR: } What city shares a border with the county where the person who started the Gold Rush in the US city having area code 951 worked in? \\
\textbf{Ground Truth: } What shares a border with the city where the person who went to the state where the 951 area code is used during the gold rush works? \\
\Xhline{1pt}
\end{tabular}
\caption{\label{tab:results/case1} Examples of multi-hop questions generated based on two and four documents.}
\label{tab:example}
\end{table*}

We also experimented with various training strategies to enable E2EQR to effectively learn multi-hop questions with diverse complexities.
Table \ref{tab:results/traning_ablation} compares the ROUGE-L scores of E2EQR models trained by four different training strategies.
\begin{enumerate}
\item In standard training procedures, the model is trained uniformly across all examples, regardless of the question complexity.
Contrasting our proposed approach, this method exhibits lower performance in generating 4-hop questions.
This outcome indicates that simultaneous learning from examples of various complexity levels results in inadequate comprehension of highly complex questions.
\item In step-by-step curriculum learning, the model is optimized sequentially, starting from generating simple questions and progressing to generating complex ones.
We trained the model by configuring the hyperparameters of Algorithm \ref{algo:curriculum learning} as follows: \{$\gamma_{low}=0$, $\gamma_{high}=0$, $\rho=0$\}. 
As shown in the results, the model experiences catastrophic forgetting with respect to the 2-hop questions initially learned.
\item In cumulative curriculum learning, the model sequentially learns from simple to complex examples, akin to the previous method.
However, in this approach, the easier examples on which the model has already been optimized are also integrated into subsequent training process.
We trained the model with the setting \{$\gamma_{low}=1$, $\gamma_{high}=0$, $\rho=0$\}.
While this method facilitates effective learning of questions with diverse complexities, it is imperative to mitigate overfitting to the simpler questions.
\item In our adaptive curriculum learning, the model learns a small number of complex questions even during the initial iterations that primarily focus on easier examples.
This approach ensures that the model trains on easier examples while also considering the necessity for rewriting to generate more complex questions.
As a result, our method empowers the model to achieve robust and balanced performance across multiple complexity levels.
\end{enumerate}

\subsection{Case Study}
Table \ref{tab:results/case1} shows questions generated by different multi-hop QG models based on the same context documents.
\textit{Example 1} illustrates the case of generating a 2-hop question based on two given documents.
The questions generated by BART and E2EQR suggest correct reasoning processes that ultimately lead to the input answer.

\textit{Example 2} depicts a more challenging scenario as the models are required to generate 4-hop questions that encompass the content of four distinct documents.
BART generates questions by using key entities present in \textit{Document B} and \textit{Document D}.
Given the question, \textit{``the area code 951''} can be replaced with \textit{``California''} based on information from \textit{Document B}, while \textit{``where the California Gold Rush began''} can be substituted with \textit{``San Francisco''}, as indicated in \textit{Document D}.
As a consequence of question simplification, the question generated by BART is equivalent to \textit{``What is the capital of California of San Francisco?''}, which is neither logical nor consistent with the answer.

In contrast, the question generated by E2EQR facilitates sequential simplification based on the given documents. 
\textit{``What city shares a border with the county where the person who started the Gold Rush in California worked in?''} transforms into \textit{``What city shares a border with the county where Samuel Brannan worked in?''}, further simplifying to \textit{``What city shares a border with Sacramento?''}
It can be confirmed that the finally obtained 1-hop question matches the input answer.

\begin{table}[h]
\centering
\scriptsize
\renewcommand*{\arraystretch}{1.2}
\begin{tabular}{m{7cm}}
\Xhline{1pt}
\textbf{Answer}: \textcolor{myyellow}{July 22, 1864} \\
\hline
\textbf{[Document A] Battle of Atlanta} \\
\textcolor{mypink}{The Battle of Atlanta } was a battle of the Atlanta Campaign fought during the American Civil War on \textcolor{myyellow}{July 22, 1864}, just southeast of Atlanta, Georgia. Continuing their summer campaign to seize the important rail and supply center of Atlanta, Union forces commanded by William Tecumseh Sherman overwhelmed and defeated Confederate forces defending the city under John Bell Hood. … \\
\textbf{Intermediate Question (1-hop)}: When did \textcolor{mypink}{the battle of Atlanta} happen? \\
\hline
\textbf{[Document B] List of municipalities in Georgia} \\
The largest municipality by population in \textcolor{myblue}{Georgia} is \textcolor{mypink}{Atlanta} with 420,003 residents, and the smallest municipality by population is Edge Hill with 24 residents. … \\
\textbf{Intermediate Question (2-hop)}: When did the Battle of \textcolor{mypink}{the largest municipality in} \textcolor{myblue}{Georgia} happen? \\
\hline
\textbf{[Document C] WEKL} \\
WEKL, known on-air as "K-Love", is a Contemporary Christian radio station in the United States, licensed by the Federal Communications Commission (FCC) to Augusta, \textcolor{myblue}{Georgia}, broadcasting on 102.3 MHz with an ERP of 1.5 kW. … \\
\textbf{Final Question (3-hop)}: When did the Battle of \textcolor{mypink}{the largest municipality in} \textcolor{myblue}{the state WEKL broadcasts in} happen? \\
\Xhline{1pt}
\end{tabular}
\caption{\label{tab:result/case2} An example of question rewriting in E2EQR. The expressions referring to the same object in the same color.}
\end{table}

Table \ref{tab:result/case2} presents an example of question rewriting using E2EQR.
The 1-hop question was generated based on \textit{Document A} and contains the bridge entity \textit{``Atlanta''} that connects \textit{Document A} and \textit{B}.
Next, our model rewrote the question to a 2-hop question using the additional information about \textit{``Atlanta''} in \textit{Document B} and included the bridge entity \textit{``Georgia''} connecting \textit{Document B} and \textit{C}.
In the final step, \textit{``Georgia''} in the previous question was replaced with the information about \textit{``WEKL''} appeared in \textit{Document C}.

\section{Conclusion}
This paper introduces a novel multi-hop question generation model that sequentially enhances question complexity based on the input documents.
Employing a step-by-step question rewriting strategy, the proposed model generates multi-hop questions, yet remains trainable end-to-end without reliance on labeled intermediate questions.
Experimental results confirm the model's effectiveness in generating complex questions requiring inference across multiple documents, while accurately aligning with input answers.
Furthermore, the synthetic QA pairs generated by our model are also beneficial when used as training data for multi-hop QA models.

\section{Ethics Statement}
This study utilized commonly used QA datasets devoid of ethical concerns. 
Furthermore, we ensured ethical practices by providing fair compensation to human evaluators recruited from Amazon Mechanical Turk and Upwork.

\section{Acknowledgements}
This work was supported by Institute of Information \& communications Technology Planning \& Evaluation (IITP) grant funded by the Korea government(MSIT) (No.2022-0-00653, Voice Phishing Information Collection and Processing and Development of a Big Data Investigation Support System)
This work was also supported by Institute of Information \& communications Technology Planning \& Evaluation (IITP) grant funded by the Korea government(MSIT) (No.2019-0-01906, Artificial Intelligence Graduate School Program(POSTECH)).

\nocite{*}
\section{Bibliographical References}\label{sec:reference}
\bibliographystyle{lrec-coling2024-natbib}
\bibliography{bibliographical_references}


\appendix

\section{\label{sec:appendix/llm} LLM-based Multi-hop Question Generation}

Recently, large language models (LLMs) have demonstrated remarkable performance across various NLP tasks.
We also investigated whether LLMs are capable of generating multi-hop questions through in-context learning.
In this experiment, we assign two tasks to GPT-3.5\footnote{\url{https://platform.openai.com/}}: standard multi-hop QG, where the model generates multi-hop questions based on the given documents and answers, and incremental multi-hop QG, wherein the model generates questions sequentially from 1-hop to the target complexity level.
In the incremental generation, we require the model to specify the index of the referenced document to generate or revise the question.
The prompt templates are described in Table \ref{tab:3shot_template} and \ref{tab:cot_template}.

\begin{table}[h]
\centering
\scriptsize
\renewcommand*{\arraystretch}{1.2}
\begin{tabular}{m{7cm}}
\Xhline{1pt}
Generate a multi-hop question for the given answer which requires reference to all of the given documents. \\
(Example1) \\
Document1: \{\{document1\}\} \\
Document2: \{\{document2\}\} \\
Document3: \{\{document3\}\} \\
Answer: \{\{answer\}\} \\
Question: \{\{question\}\} \\ \\
... \\ \\
(Example4) \\
Document1: \{\{document1\}\} \\
Document2: \{\{document2\}\} \\
Document3: \{\{document3\}\} \\
Answer: \{\{answer\}\} \\
Question: \\
\Xhline{1pt}
\end{tabular}
\caption{\label{tab:3shot_template} Prompt template used for standard generation. We attached three exemplars randomly selected from the MuSiQue training set.}
\end{table}

\begin{table}[h]
\centering
\scriptsize
\renewcommand*{\arraystretch}{1.2}
\begin{tabular}{m{7cm}}
\Xhline{1pt}
Generate a multi-hop question for the given answer which requires reference to all of the given documents. \\
(Example1) \\
Document1: \{\{document1\}\} \\
Document2: \{\{document2\}\} \\
Document3: \{\{document3\}\} \\
Answer: \{\{answer\}\} \\
1-hop question (using Document$i$): \{\{1-hop question\}\} \\
2-hop question (using Document$j$): \{\{2-hop question\}\} \\
3-hop question (using Document$k$): \{\{3-hop question\}\} \\
... \\ \\
(Example4) \\
Document1: \{\{document1\}\} \\
Document2: \{\{document2\}\} \\
Document3: \{\{document3\}\} \\
Answer: \{\{answer\}\} \\
\Xhline{1pt}
\end{tabular}
\caption{\label{tab:cot_template} Prompt template used for incremental generation. We randomly selected three exemplars from the MuSiQue training set and manually labeled their intermediate questions.}
\end{table}

\begin{table}[]
\centering
\renewcommand*{\arraystretch}{1.2}
\resizebox{\columnwidth}{!}{
\begin{tabular}{ll|ccc}
\Xhline{1pt}
\multicolumn{2}{l|}{Model}                                  & 2-hop & 3-hop & 4-hop  \\ \hline
\multicolumn{1}{l|}{\multirow{2}{*}{GPT-3.5}} & Standard    & 23.54 & 22.92 & 20.64  \\
\multicolumn{1}{l|}{}                         & Incremental & 26.04 & 25.98 & 21.86 \\ \hline
\multicolumn{2}{l|}{E2EQR}                                  & 44.01 & 40.04 & 36.98  \\ \Xhline{1pt}
\end{tabular}}
\caption{\label{tab:gpt_automatic} ROUGE-L scores of GPT-3.5 models.}
\end{table}

According to the results presented in Table \ref{tab:gpt_automatic}, GPT-3.5 exhibits inferior performance compared to E2EQR.
However, GPT-3.5 demonstrates better performance in incremental generation, wherein the model progressively enhances question complexity, than in standard generation.
This observation indicates the effectiveness of sequential question rewriting approach to generate complex questions, even for advanced LLMs.

\begin{table}[h]
\centering
\scriptsize
\renewcommand*{\arraystretch}{1.2}
\begin{tabular}{m{7cm}}
\Xhline{1pt}
\textbf{Answer}: \textit{Tamaulipas} \\
\hline
\textbf{[Document 1]}
The Municipality of Nuevo Laredo is located in the Mexican state of \textit{Tamaulipas}. ... \\
\textbf{[Document 2]}
The Ciudad Deportiva (\"Sports City\") is a sports complex in Nuevo Laredo, Mexico. It is home to the Tecolotes de Nuevo Laredo Mexican Baseball League team and ... \\
\hline
\textbf{Standard Generation} \\
Which Mexican state is home to the municipality containing Nuevo Laredo and Ciudad Deportiva? \\
\hline
\textbf{Incremental Generation} \\
1-hop question (using Document1): In which Mexican state is Nuevo Laredo located? \\
2-hop question (using Document2): In which Mexican state can one find the Ciudad Deportiva, home to the Tecolotes de Nuevo Laredo? \\
\hline
\textbf{Ground Truth} \\
What administrative territorial entity is the owner of Ciudad Deportiva located? \\
\Xhline{1pt}
\textbf{Answer}: \textit{Milledgeville} \\
\hline
\textbf{[Document 1]}
Blackberry Smoke is an American rock band from Atlanta, Georgia, United States.  ... \\
\textbf{[Document 2]}
WEKL, known on-air as "102.3 K-Love", is a Contemporary Christian radio station in the United States, ...\\
\textbf{[Document 3]}
Georgia has had five different capitals in its history. The first was Savannah, the seat of government during British colonial rule, followed by Augusta, Louisville, \textit{Milledgeville}, and Atlanta, the capital city from 1868 to the present day. ... \\
\hline
\textbf{Standard Generation} \\
Where is the location of the studios for the radio station in Augusta where the American rock band Blackberry Smoke originates from? \\
\hline
\textbf{Incremental Generation} \\
1-hop question (using Document3): What city served as the capital of Georgia before Atlanta? \\
2-hop question (using Document2): What city in Georgia is the 102.3 K-Love radio station located in? \\
3-hop question (using Document1): What city in Georgia is the band Blackberry Smoke from? \\
\hline
\textbf{Ground Truth} \\
What was the capital of the state where WEKL operates, before the city where Blackberry Smoke was formed? \\
\Xhline{1pt}
\end{tabular}
\caption{\label{tab:gpt_examples} Examples of 2- and 3-hop questions generated by GPT-3.5.}
\end{table}

We analyzed the prevalent error cases in the generation results of GPT-3.5.
Table \ref{tab:gpt_examples} illustrates representative examples of 2- and 3-hop generated questions.
In the standard generation, the model can be seen to successfully generates questions that incorporate content derived from multiple documents.
However, the model tends to produce questions that can be resolved by referring to a single document.
For example, the 2-hop question generated in standard setting includes \textit{``Ciuda Deportiva''} from \textit{Document 2}, yet can be answered by reading only \textit{Document 1}.
Similar outcomes are observed in 3-hop QG, where the question does not even match the input answer.
We believe that while GPT-3.5 recognizes and mimic patterns from few-shot examples, it encounters limitations in logically generating multi-hop questions that correspond appropriately with the input answers.

Furthermore, during incremental generation, GPT-3.5 encounters difficulties in generating 3- and 4-hop questions, frequently formulating the questions that primarily focus on the content of the last referenced document.
As shown in the 3-hop generation, the model inserts information from a newly referenced document into the question without retaining the main content of the previous question.
In conclusion, in-context learning is not sufficient to achieve the objective of generating logically structured multi-hop questions that match the input answers, and our model is superior at multi-hop QG tasks although it requires supervised learning.

\section{\label{sec:appendix/hotpotqa} Automatic Evaluation on HotpotQA}
\begin{table}[h]
\centering
\renewcommand*{\arraystretch}{1.2}
\resizebox{\columnwidth}{!}{
\begin{tabular}{l|ccc}
\Xhline{1pt}
Model                           & BLEU-4 & METEOR & ROUGE-L \\ \hline
DP-Graph \cite{pan2020semantic} & 15.53 & 20.15  & 36.94   \\
MulQG \cite{su2020multi}        & 15.20 & 20.51  & 35.30   \\
CQG \cite{fei2022cqg}           & 21.46 & 24.97  & 39.61   \\
BART \cite{lewis2020bart}       & 21.77 & 29.40  & 43.01   \\
E2EQR                           & 21.73 & 28.47  & 41.34   \\ \Xhline{1pt}
\end{tabular}}
\caption{\label{tab:hotpot-auto} Automatic evaluation results on HotpotQA.}
\end{table}
We compared the performance of various multi-hop QG models on HotpotQA, which contains two question types: \textit{bridge} and \textit{comparison}.
Because our model cannot generate the comparison question using question rewriting, we handled these as 1-hop questions.
As presented in Table \ref{tab:hotpot-auto}, the proposed model achieves slightly lower performance than BART but outperforms the remaining baselines.
It seems that BART performs better than E2EQR because the supporting facts in the HotpotQA dataset are shorter than those from the MuSiQue dataset, and two-hop QG is sufficiently possible with a seq2seq model.

\end{document}